\pdfoutput=1

\documentclass[11pt]{article}

\usepackage[final]{acl}

\usepackage{times}
\usepackage{latexsym}

\usepackage[T1]{fontenc}

\usepackage[utf8]{inputenc}

\usepackage{microtype}

\usepackage{inconsolata}

\usepackage{graphicx}

\usepackage{multirow}
\usepackage{subfig}
\usepackage{booktabs}
\usepackage{colortbl} 

\usepackage{listings}
\usepackage{xcolor}
\definecolor{lightgray}{gray}{0.9}

\definecolor{clipcomment}{RGB}{70,130,180} 

\lstdefinelanguage{PseudoPython}{
    language=Python,
    morekeywords={np, arange, dot, exp, cross_entropy_loss, l2_normalize},
    sensitive=true
}

\lstdefinestyle{clipstyle}{
  language=PseudoPython,
  basicstyle=\ttfamily\small,
  commentstyle=\color{clipcomment},
  keywordstyle=\color{black},
  stringstyle=\color{black},
  backgroundcolor=\color{white},
  showstringspaces=false,
  columns=fullflexible,
  keepspaces=true,
  frame=none,
  numbers=none,
  escapeinside={(*@}{@*)},
  literate={§}{{\S}}1
}

\usepackage{amssymb}
\usepackage{amsmath} 
\usepackage{makecell}

\usepackage{siunitx}

\title{Segment, Embed, and Align:\\A Universal Recipe for Aligning Subtitles to Signing\\}

\author{
\parbox{0.8\linewidth}{\centering
Zifan Jiang$^{1,2}$*, Youngjoon Jang$^{1,3}$, Liliane Momeni$^{1}$, \\
Gül Varol$^{1,4}$, Sarah Ebling$^{2}$, Andrew Zisserman$^{1}$,}
\\
$^1$VGG, Dept.\ of Engineering Science, University of Oxford, 
$^2$University of Zurich, 
\\
$^3$KAIST, 
$^4$LIGM, CNRS, Univ Gustave Eiffel, ENPC, IP Paris
\\
\texttt{jiang@cl.uzh.ch} \\
\url{https://j22melody.github.io/SEA/}
}

\begin{document}
\maketitle
\def\thefootnote{*}\footnotetext{Work done during a research visit to VGG, Oxford.}
\def\thefootnote{\arabic{footnote}}

\begin{figure*}
    \centering
    \includegraphics[width=\linewidth]{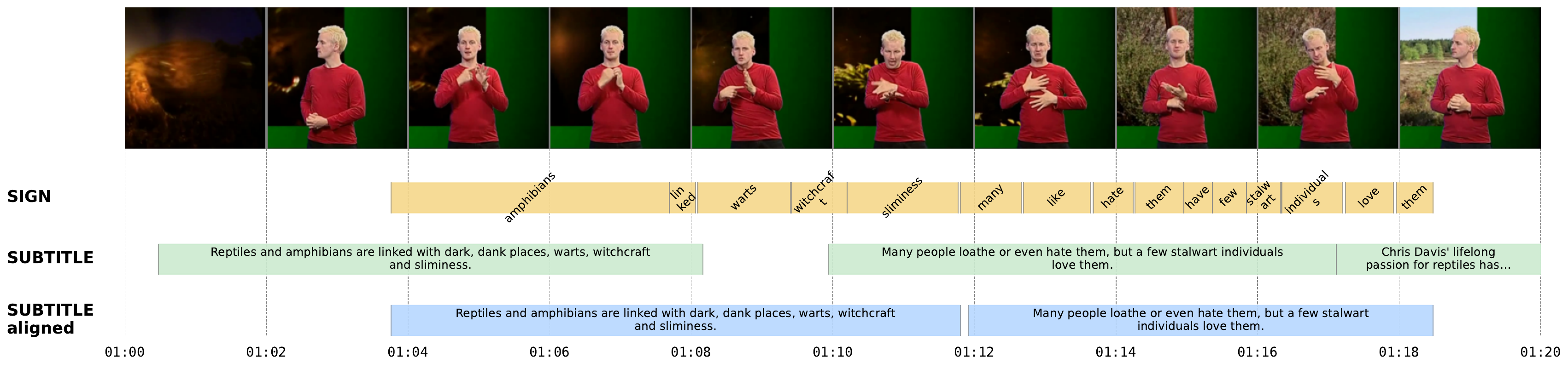}
    \caption{\textbf{Aligning subtitles to continuous signing:} In broadcast interpreting and other sign language corpora, original subtitles (green) frequently lag or lead the actual signing (yellow) by non-deterministic amounts. Our alignment method, SEA, produces time-corrected subtitles (blue) that better correspond to the signed content. Keyframes are sampled at the midpoint of each span and may not include all annotated signs present in that interval.}
    \label{fig:teaser}
\end{figure*}

\begin{abstract}
The goal of this work is to develop a universal approach for aligning subtitles (i.e., spoken language text with corresponding timestamps) to continuous sign language videos. Prior approaches typically rely on end-to-end training tied to a specific language or dataset, which limits their generality. In contrast, our method Segment, Embed, and Align (SEA) provides a single framework that works across multiple languages and domains. SEA leverages two pretrained models: the first to segment a video frame sequence into individual signs and the second to embed the video clip of each sign into a shared latent space with text. Alignment is subsequently performed with a lightweight dynamic programming procedure that runs efficiently on CPUs within a minute, even for hour-long episodes. SEA is flexible and can adapt to a wide range of scenarios, utilizing resources from small lexicons to large continuous corpora. Experiments on four sign language datasets demonstrate state-of-the-art alignment performance, highlighting the potential of SEA to generate high-quality parallel data for advancing sign language processing.
SEA's code and models are openly available\footnote{\url{https://github.com/J22Melody/SEA}}.

\end{abstract}

\section{Introduction}

Sign language alignment is the task of aligning spoken language text to continuous sign language videos. 
Formally, given a video and a sequence of weakly temporally aligned subtitle units, the alignment task adjusts the start/end timestamps of the subtitles to better match the signing (Figure~\ref{fig:teaser}). This matters in practice for at least two reasons.

First, the development of sign language translation systems \citep{muller-etal-2022-findings,muller-etal-2023-findings,de2023machine,ham2021ksl} relies on \textit{parallel data} between sign language videos and spoken language text. While small-scale studio or lab datasets, such as How2Sign \citep{Duarte_CVPR2021} and CSL-Daily \citep{zhou2021improving}, offer high-quality manual alignment, large-scale broadcast datasets, like BOBSL \citep{dataset:albanie2021bobsl}, often exhibit noisy alignment. Subtitles for broadcast content are typically generated based on the original audio track and therefore show a non-deterministic temporal offset relative to the signed interpretation \citep{muller-etal-2022-findings}. Improving alignment quality benefits the training and evaluation of translation models.


Second, user-generated sign language content on online platforms like YouTube \citep{uthus_youtube-asl_2023,tanzer2025youtubesl25} gains from \emph{automatic subtitling}.
Good alignment cuts the manual effort of post-hoc captioning and improves accessibility at scale \citep{gernsbacher2015video}. 
Accurate alignment also helps sign language linguists when creating and annotating corpora—e.g., the DGS Corpus \citep{dataset:hanke-etal-2020-extending}—by pre-segmenting material and synchronizing text and signing tiers.

The most reliable way to align is through human annotators (ideally, native signers) with tools such as ELAN \cite{wittenburg2006elan}, VIA \cite{dutta2019vgg}, or iLex \cite{hanke-2002-ilex}.
However, manual alignment is prohibitively time-consuming and costly.
As reported by \citet{bull2021aligning} in the context of BOBSL, an expert fluent in sign language takes approximately 10–15 hours to align subtitles to one hour of continuous sign language video. 
On the other hand, the per-hour rate is $\sim$40 USD according to the WMT-SLT 22 campaign \cite{muller-etal-2022-findings}.
Previous methods for sign language alignment have either relied directly on manually aligned data or indirectly on an end-to-end network trained on the former. 
For example, \citet{bull2021aligning} finetunes a subtitle aligner network on 17.7 hours of manual annotation (\S\ref{sec:related}).

In this work, we propose \textit{Segment, Embed, and Align} (SEA) as a general solution to the text-signing alignment problem, which is effective across various languages and data sources and requires almost zero direct in-domain supervision for the alignment task. 
SEA consists of three modular steps:
(1) \textit{Segment} continuous sign language video frames into individual meaningful units (e.g., signs);
(2) \textit{Embed} the sequences of signing and subtitle units of a video into a shared latent space;
(3) \textit{Align} the subtitle units to the signs based on a similarity matrix formed by text and signing embeddings and the prior subtitle locations.
We highlight that the first two modules are facilitated by pretrained sign language  \textit{segmentation} \citep{moryossef-etal-2023-linguistically} and \textit{embedding} \citep{jiang-etal-2024-signclip} models that transfer to alignment as a downstream task. 
The alignment module is a lightweight {\em global} optimization based on dynamic programming (DP). 
It runs swiftly on CPUs at the episode level without requiring gradient-based operations. 
In contrast, prior work performs training/inference in {\em local} windows of about 20 seconds, followed by a slow dynamic time warping (DTW) process to resolve local conflicts \citep{bull2021aligning}.

SEA’s modular design offers greater generalisability than end-to-end, in-domain supervised methods: individual components can be seamlessly swapped in a plug-and-play manner for different languages while keeping the framework unchanged. Our experiments (\S\ref{sec:exp}) reveal that the segmentation component, in particular, exhibits relatively strong cross-linguistic universality, while the embedding component benefits from language-specific expertise. 
SEA achieves state-of-the-art (SOTA) performance across four different benchmarks (BOBSL, How2Sign, WMT-SLT, and SwissSLi \cite{jiang-etal-2024-swisssli}) and three sign languages: British Sign Language (BSL), American Sign Language (ASL), and Swiss German Sign Language (DSGS).
All SEA modules are open for the curation of better-aligned sign language datasets and future improvements.


\begin{figure*}[ht]
    \centering
    \includegraphics[width=\linewidth]{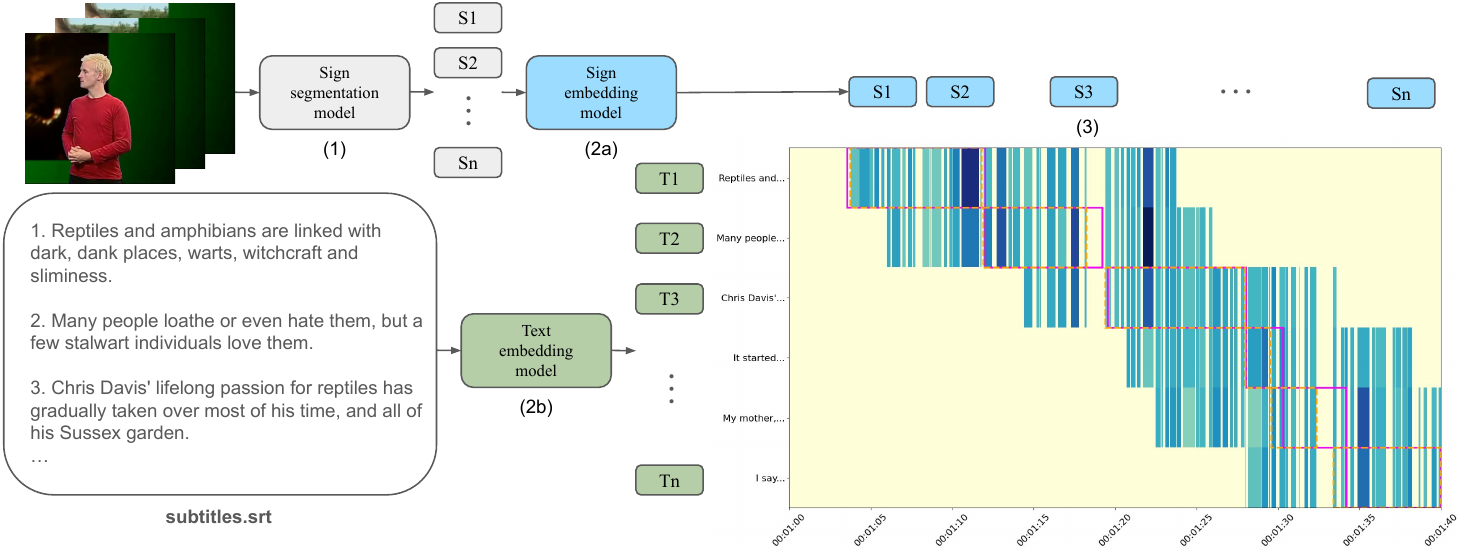}
    \caption{\textbf{Our method SEA consists of three modular steps:} (1) \textit{Segment} video frames of continuous signing into individual signs; (2) \textit{Embedd} each sign (\texttt{$s_1$} to \texttt{$s_n$}; 2a) and subtitle unit (\texttt{$t_1$} to \texttt{$t_n$}; 2b) into a shared latent space, with their dot product similarities encoded by a similarity matrix; (3) \textit{Align} subtitles to signing based on the text-sign similarities and the original temporal location of the subtitle units. The similarity matrix is illustrated as a heatmap over time, with darker bars indicating a higher similarity between a sign and a subtitle; similarities of the originally remote signs are zeroed out to ensure locality. The dashed/solid boxes in the heatmap indicate the predicted/manually aligned subtitle locations, respectively.}
    \label{fig:method}
\end{figure*}

\section{Related Work}
\label{sec:related}






\paragraph{Video-text alignment.} 

General video–text alignment has been widely studied on HowTo100M \citep{miech19howto100m}, a large corpus of instructional videos paired with narrations derived from automatic speech recognition, which provides weak supervision. 
Building on this, \citet{miech20endtoend} proposes MIL-NCE to learn joint video–text representations through multi-instance contrastive learning, addressing narration–visual misalignment. 
\citet{han2022temporal} further uses weak labels to estimate sentence alignability and predict temporal spans, enabling sentence-level localization in long videos.

\paragraph{Signing-subtitle alignment.}

Specifically developed for sign language videos, \citet{bull2021aligning} introduces a model named Subtitle Aligner Transformer (SAT) that is first pretrained on large-scale data for the task of sign localisation with weak supervision and then finetuned on 17.7 hours of manually aligned subtitle-signing data \cite{albanie2020bsl}. 
SAT utilizes BERT \cite{devlin-etal-2019-bert} to embed the subtitles and I3D \cite{carreira2017quo} to embed the visual signal, then cross-attends them in an encoder-decoder architecture. 
The model outputs frame-level predictions on whether each frame belongs to the queried subtitle or not within a 20-second search window. 

A contemporary study, \citet{jang2025deep}, extends SAT to a new SOTA on BOBSL by three additional contributions: 
(1) Preprocessing the subtitles with specific grammatical
rules;
(2) A selective alignment loss with negative sampling of irrelevant video clips;
(3) Iterative self-training with refined pseudo-labels.
These modifications optimize SAT, but in contrast to SEA, a fair amount of manually aligned data remains key to achieving strong performance. 
Furthermore, although SAT is end-to-end at the sentence level, \citet{bull2021aligning,jang2025deep} must follow local model predictions using a DTW process to resolve the temporal conflicts of individual predictions. On the other hand, the alignment in the SEA algorithm runs globally at the episode level, so no local window is needed. 

\citet{tanzer-2025-fleurs} proposes a multitasking network with two branches, caption alignment and translation, inspired by Whisper \cite{radford2023robust}.
However, the reported alignment performance, measured by frame accuracy, lags far behind a model-free baseline where the caption duration is scaled in proportion to the caption length.
We present SEA as a training-free alternative for the alignment task, as opposed to prior work.

\section{Method}
\label{sec:method}
In this section, we describe the SEA method (Figure~\ref{fig:method}) step by step.
We first introduce how we leverage pretrained sign language \textit{segmentation} (\S\ref{sec:segmentation}) and \textit{embedding} (\S\ref{sec:embedding}) models, and then present the key \textit{alignment} step (\S\ref{sec:alignment}).

\subsection{Segmentation}
\label{sec:segmentation}

Given a continuous stream of video frames, also known as an episode, we first segment it into \textit{meaningful units}, such as signs\footnote{Practically, the signing units can be on a sub-sign level due to an imperfect segmentation model.}. 
This process is analogous to text tokenization.
While the idea sounds natural, in practice, there is no standardized tool like those in NLP \cite{sennrich-etal-2016-neural,kudo-richardson-2018-sentencepiece}, and the sign boundaries are not clearly defined linguistically \cite{hanke2012does}. 

In this work, we adopt the automatic sign segmentation model of \citet{moryossef-etal-2023-linguistically}, trained on $\sim$73 hours of annotated data from the Public DGS Corpus. 
It passes MediaPipe Holistic \cite{mediapipe2020holistic} poses into a lightweight LSTM \cite{hochreiter1997long} model and then produces frame-level predictions.
Despite being trained only on German Sign Language (DGS), the model shows successful zero-shot transfer to other sign languages\footnote{Zero-shot transfer (0.76 ROC-AUC on \textit{O}-tag) was reported for French Sign Language in \citet{moryossef-etal-2023-linguistically}, and we find the segmentation also useful for ASL/BSL/DSGS.}, which is attributed to (1) the iconicity and finite set of hand/body configurations in sign languages, and (2) the use of MediaPipe pose representations that capture and focus on these motion primitives. 


\subsection{Embedding}
\label{sec:embedding}

After sign segmentation, for each episode, we now have a sequence of \textit{signs} and a sequence of \textit{subtitle units}.
Each \textit{sign} is a sequence of continuous video frames of signing, and each \textit{subtitle} is a sequence of text tokens\footnote{Subtitles may or may not form well-formed sentences.}.
The next goal is to convert them into sequences of \textit{alignable vectors}. 

We employ the SignCLIP \cite{jiang-etal-2024-signclip} model, which re-purposes CLIP \cite{radford2021learning} to
project text and signing into the same latent space.
Subtitle units are embedded with a BERT-like text encoder, while signing video clips are converted to a sequence of MediaPipe poses and then encoded by a pose encoder initialized from the same BERT weights and finetuned for signing. 
We use a multilingual checkpoint pretrained on \textit{\href{https://www.spreadthesign.com/}{SpreadtheSign}} by \citet{jiang-etal-2024-signclip}, which covers 41 sign languages and English on the text side only.
We also show the benefits of language-specific finetuning for SignCLIP on BSL/English, ASL/English, and DSGS/German data in \S\ref{sec:exp}.

\subsection{Alignment}
\label{sec:alignment}

The next goal is to assign each subtitle to a contiguous span of predicted sign segments, and then rewrite the subtitle’s timestamps to reflect that span. 
We optimize a single objective that balances two kinds of evidence. 
(1) Temporal/prosodic costs: how far the subtitle must move, how much the duration changes, and how big the inter-sign gaps are within a subtitle—signing tends to be continuous, and big gaps indicate prosodic pauses that lead to the next phrase. 
(2) Semantic similarity costs: a reward for assigning the subtitle to signs whose embeddings are similar to the text. 
We consider two variants within the same framework: \textit{Segment and Align} uses only the timing/prosodic costs (no semantics), while \textit{Segment, Embed, and Align} augments the same objective with the semantic term. 
Both variants are solved by a single global DP procedure over the full episode.
We mathematically define the alignment algorithm next, present implementation details in \S\ref{sec:implementation}, and provide a pseudocode description in Appendix \S\ref{sec:extended_details} for additional clarity.

\paragraph{Task.} Given subtitle units $t_1,\dots,t_n$ with times $(\mathrm{start}(t_i),\mathrm{end}(t_i))$ and signs $s_1,\dots,s_m$ with $(\mathrm{start}(s_j),\mathrm{end}(s_j))$, we assign to each $t_i$ a contiguous span ${s_l,\dots,s_r}$ and shift $t_i$ to that span’s boundaries $(\mathrm{start}(s_l),\mathrm{end}(s_r))$. 

\paragraph{Cost function.}
For a pair $(t_i,{s_l,\dots,s_r})$, a \textit{cost function} $\phi(t_i,s_{l{:}r})$ is defined by a weighted sum of: onset distance $|\mathrm{start}(t_i)-\mathrm{start}(s_l)|$, offset distance $|\mathrm{end}(t_i)-\mathrm{end}(s_r)|$, duration difference $|(\mathrm{end}(t_i)-\mathrm{start}(t_i))-(\mathrm{end}(s_r)-\mathrm{start}(s_l))|$ with weight $w_{\mathrm{dur}}$, and internal inter-sign gaps $G(l,r){=}\sum_{j=l}^{r-1}\max(0,\mathrm{start}(s_{j+1})-\mathrm{end}(s_j))$ with weight $w_{\mathrm{gap}}$.

\paragraph{Segment and Align.}
We select non-overlapping spans $(l_i,r_i)$ for $t_i$ that minimize the total cost $\sum_i \phi(t_i,s_{l_i{:}r_i})$ using the individual costs above.
A standard episode-level DP algorithm is employed to determine the global optimum, which iterates over the subtitles and signs once their matching costs are set up.
The DP is similar to, but distinct from, DTW: DTW treats each unit independently, except for path continuity; whereas SEA considers inter-sign gaps as part of the cost.
We append the formal definitions of the DP in Appendix \S\ref{sec:extended_details}.

\paragraph{Segment, Embed, and Align.}
We enrich alignment with a text–sign similarity matrix. Let $u_i$ and $v_j$ be embeddings for $t_i$ and $s_j$; define raw similarities by dot product $M^0_{ij}{=}\langle u_i,v_j\rangle$. For each row $i$, we keep only $window\_size=50$ temporally nearest signs to a subtitle to ensure locality (others masked to $0$) and apply a row-wise \textit{softmax} normalization to obtain the final $M$. 
The total span similarity is $\Sigma(i;l,r){=}\sum_{j=l}^{r} M_{ij}$, and we add a negative term $-w_{\mathrm{sim}}\Sigma(i;l,r)$ to the cost function.
It pushes the algorithm to align subtitles to a span of similar signs based on the weight $w_{\mathrm{sim}}$.
The DP optimization process is identical to the above.

\begin{table*}[ht]
\centering
\resizebox{\linewidth}{!}{%
\begin{tabular}{lllrrrrrr}
\toprule
\textbf{Dataset} & \textbf{Languages} & \textbf{Data source} & \textbf{\#episodes (train/dev/test)} & \textbf{\#subtitles} & \textbf{\#hours} & \textbf{\#signers} & \textbf{Offsets} & \textbf{FPS} \\ 

\midrule

BOBSL & BSL/English & Interpretation of mixed programs & 55 (16/4/35) & 31,479 & $\sim$46 & 22 & 2.70/2.70 & 25 \\

How2Sign & ASL/English & Studio-recorded re-signing & 2,528 (2,212/132/184) & 35,191 & $\sim$80 & 9 & 1.57/2.13 & 24/30/50/60 \\

WMT-SLT SRF & DSGS/German & Live interpretation of daily news & 29 (21/4/4) & 7,071 & $\sim$16 & 3 & 1.06/1.75 & 25 \\

SwissSLi mitenand & DSGS/German & Offline, deaf translation of one program & 14 (0/7/7) & 554 & $\sim$1 & 2 & 0.49/0.56 & 25  \\

\bottomrule
\end{tabular}
}
\caption{Summary of datasets and their statistics. \textit{Offsets} corresponds to the temporal shifts added to the start/end of the subtitles in the \textit{Original\textsuperscript{+}} baseline, where the offsets correspond to the median (or mean) differences between the original and manually aligned subtitle timings, calculated from the training (or validation for SwissSLi) set.}
\label{tab:datasets}
\end{table*}

\subsection{Implementation Details}
\label{sec:implementation}

\paragraph{Segmentation.} We use the \texttt{E4s-1} checkpoint from \citet{moryossef-etal-2023-linguistically} as the sign segmentation model, with two controllable decoding parameters \texttt{b-threshold} and \texttt{o-threshold} ranging from 0 to 100 to decide when to start/end a sign.

For BSL, we explore finetuning the segmentation model on 3.3 hours (149 videos) of the BSL Corpus \cite{schembri2013building}, a video corpus of conversational BSL with gloss-level annotations. 
Given the small size, we use \textit{negative sampling}: for each annotated sentence, a non-signing clip from BOBSL (10+ seconds away from any detected sign, lasting 1–30s) is randomly prepended, appended, or substituted. 
It helps the model to recognize negatives, as the data is densely packed with signing; the best model is named \textit{Segmentation-BSLCP}.

We also evaluate the BSL Corpus–trained models of \citet{Renz2021signsegmentation_a,Renz2021signsegmentation_b} in \S\ref{sec:ablation}.
With less supervision and a focus on boundary detection rather than individual signs, they often produce false positives on non-signing segments.
In contrast, \citet{moryossef-etal-2023-linguistically} addresses it through a BIO tagging scheme \cite{ramshaw1999text}, which detects active signing.
We omit experimenting with the recent follow-up of \citet{he2025hands} as neither the model nor the code has been available.

\paragraph{Embedding.} We use the \textit{SignCLIP-multilingual} (\texttt{E7.2}; \texttt{baseline\_temporal}) checkpoint for sign and subtitle embeddings by default. 
ISO 639-3 language codes are prepended in the text prompts for aligning different language pairs: \texttt{<en> <bfi>} for BSL/English; \texttt{<en> <ase>} for ASL/English; and \texttt{<de> <sgg>} for DSGS/German.

We further finetune \textit{SignCLIP-multilingual} with language-specific data. 
For BSL/English, we use sign spottings from BOBSL \cite{Momeni22}, following \citet{raude2024tale,jang2025lost} but with a contrastive text objective instead of classification over 8,697 signs, totaling $\sim$3.5M examples. 
For ASL/English, we aggregate $\sim$200K examples from PopSign \cite{Starner2023PopSignAV}, ASL Citizen \cite{Desai2023ASLCA}, and Sem-Lex \cite{Kezar2023TheSB}, three isolated sign language recognition (ISLR) datasets. 
We normalize text by using lowercase glosses without variations. 
For DSGS/German, we finetune on Signsuisse \cite{muller-etal-2023-findings}, which provides 16,213 lexical items each paired with an example sentence. 
The resulting models are named \textit{SignCLIP-BSL}, \textit{SignCLIP-ASL}, and \textit{SignCLIP-Suisse}, respectively\footnote{The checkpoints are available at \url{https://github.com/J22Melody/fairseq/tree/main/examples/MMPT\#demo-and-model-weights}.}.
Finetuning takes less than 3 days on an NVIDIA RTX A6000 GPU with 48 GB of GPU memory.

\paragraph{Alignment.}
We start by experimenting with \textit{Segment and Align} without \textit{Embed}. 
For datasets where a training set is available, we conduct a random search to identify the optimal parameter combination, including the DP weights and parameters, the segmentation model decoding parameters, and fixed offset values applied before and after the DP process.
This search process learns dataset-specific bias, such as the common temporal delay in interpretation/translation, as summarized in Table \ref{tab:datasets}.
We then incorporate the similarity matrix provided by the embeddings with the weight $w_{\mathrm{sim}}$, which reflects the strength of the embeddings.
Additional details and the best parameter combinations are listed in Appendix \ref{sec:extended_details}.

\section{Experiments}\label{sec:exp}

\begin{table*}
    \centering
    \small
    \resizebox{\linewidth}{!}{%
    \begin{tabular}{l cc cc cc cc}
      \toprule
      \multicolumn{1}{c}{} &
      \multicolumn{2}{c}{\textbf{BOBSL}} &
      \multicolumn{2}{c}{\textbf{How2Sign}} &
      \multicolumn{2}{c}{\textbf{WMT-SLT}} &
      \multicolumn{2}{c}{\textbf{SwissSLi}} \\
      \cmidrule(lr){2-3} \cmidrule(lr){4-5} \cmidrule(lr){6-7} \cmidrule(lr){8-9}
      \multicolumn{1}{c}{} &
      \textbf{Val} & \textbf{Test} &
      \textbf{Val} & \textbf{Test} &
      \textbf{Val} & \textbf{Test} &
      \textbf{Val} & \textbf{Test} \\
      \midrule
      \rowcolor{lightgray} \multicolumn{9}{l}{\textsc{Static Baselines}} \\
      \noalign{\vskip 1pt}
Original alignment (usually audio-based) & 29.09  & 14.11 & 30.63 & 33.06 & 47.83 & 46.85 & 69.58 & 60.48 \\
Original\textsuperscript{+} (fixed offsets from training data) & 49.63 & 44.61 & 31.91 & 36.21 & 74.17 & 74.83 & -- & -- \\
\midrule
\rowcolor{lightgray} \multicolumn{9}{l}{\textsc{End-to-end Trained Baselines}} \\
\noalign{\vskip 1pt}
SAT \cite{bull2021aligning} & 60.98 & 54.57 & -- & -- & 75.32 & 75.32 & -- & --  \\
SAT\textsuperscript{+} \cite{jang2025deep} & 74.30 & 63.81 & -- & -- & -- & -- & -- & -- \\
      \midrule
      \rowcolor{lightgray} \multicolumn{9}{l}{\textsc{Segment and Align}} \\
      \noalign{\vskip 1pt}
Segmentation \cite{moryossef-etal-2023-linguistically} & 66.24 & 49.58 & 33.38 & 36.17 & \textbf{75.66} & 76.83 & 71.48 & 84.19 \\
Segmentation-BSLCP & 63.61 & 47.75 & -- & -- & -- & -- & -- & -- \\
      \midrule
      \rowcolor{lightgray} \multicolumn{9}{l}{\textsc{Segment, Embed, and Align}} \\
      \noalign{\vskip 1pt}
SignCLIP-multilingual & 66.70 & 50.68 & 35.51 & 37.51 & 75.16 & 76.43 & \textbf{71.86} & \textbf{85.57} \\
SignCLIP-BSL/ASL/DSGS-finetuned & 72.78 & 54.50 & \textbf{38.32} & \textbf{39.57} & 75.34 & \textbf{77.69} & \textbf{71.86} & 85.22  \\
SignCLIP-BSL + SAT\textsuperscript{+} subtitles & \textbf{75.27} & \textbf{65.81} & -- & -- & -- & -- & -- & --  \\
      \bottomrule
    \end{tabular}%
    }
    \caption{SEA alignment performance (F1@0.50) on the validation and test sets across four sign language datasets. SEA is SOTA on all test sets and we report two variants: (1) \emph{Segment and Align} (DP with timing/prosodic costs only); (2) \emph{Segment, Embed, and Align} (same DP + SignCLIP text–sign similarity; multilingual or finetuned). Baselines include static offsets and SAT/SAT$^{+}$. For BOBSL, we also test SEA initialized from SAT$^{+}$ predicted subtitles.}
    \label{tab:results}
\end{table*}

\begin{figure*}
    \normalsize
    \centering
    \subfloat[BOBSL (English$\rightarrow$BSL; video id: \texttt{5224144816887051284})]{
        \includegraphics[width=\linewidth]{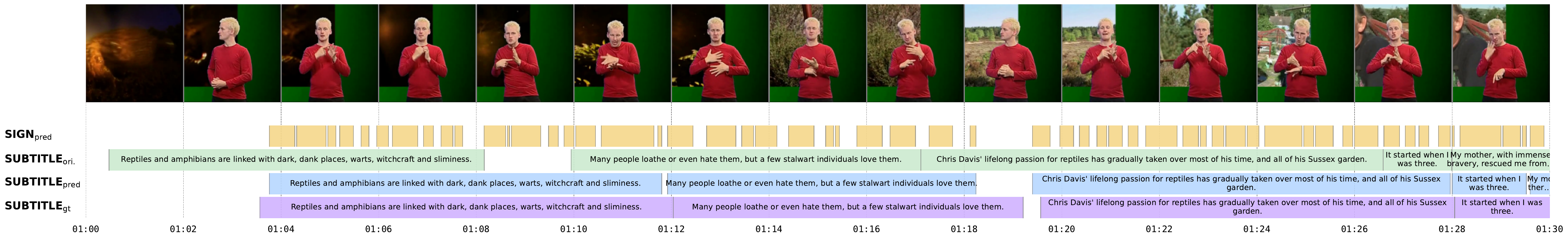}
        \label{fig:visual_bobsl}
    } \\[0.05em]
    \subfloat[How2Sign (English$\rightarrow$ASL; video id: \texttt{ETOZLBScxWY-3-rgb\_front})]{
        \includegraphics[width=\linewidth]{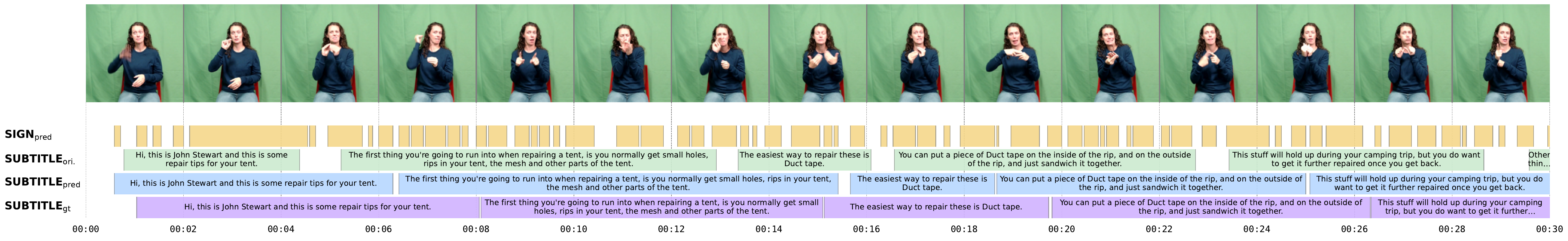}
        \label{fig:visual_how2sign}
    } \\[0.05em]
    \subfloat[WMT-SLT SRF (German$\rightarrow$DSGS; video id: \texttt{srf.2020-03-13})]{
        \includegraphics[width=\linewidth]{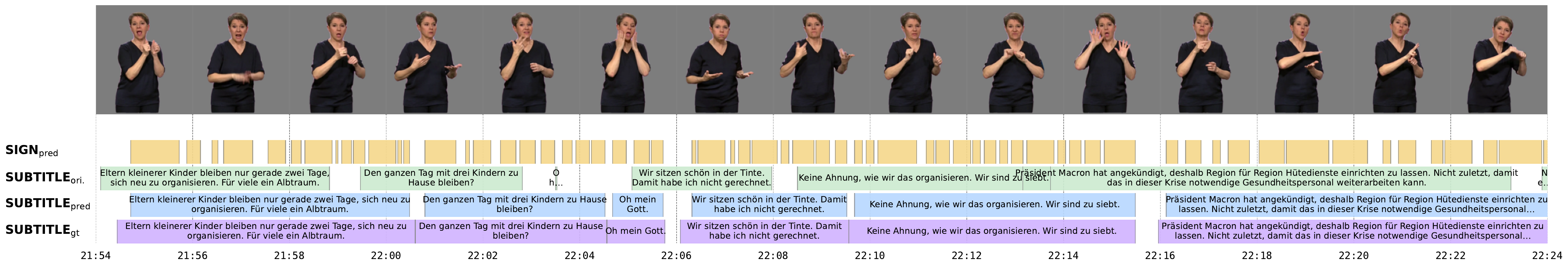}
        \label{fig:visual_wmt_slt_srf}
    } \\[0.05em]
    \subfloat[SwissSLi mitenand (German$\rightarrow$DSGS; video id: \texttt{2023-02-16\_ein\_hund\_f\"urs})]{
        \includegraphics[width=\linewidth]{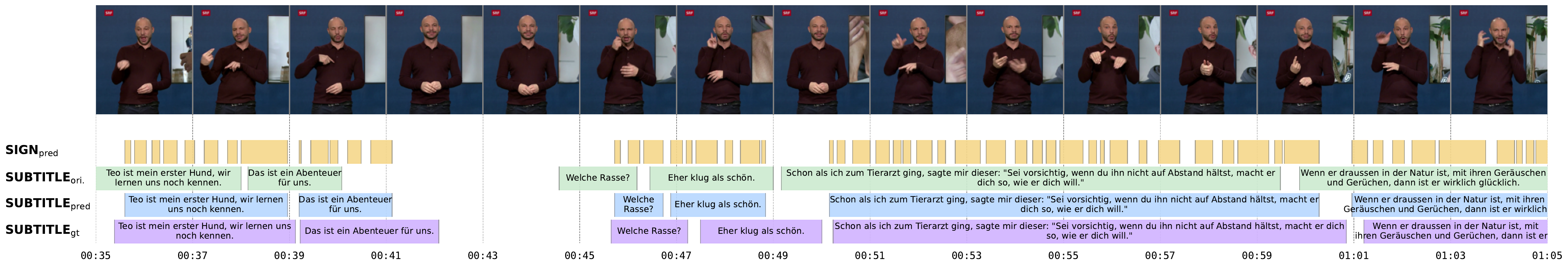}
        \label{fig:visual_swisssli}
    }
    \caption{Qualitative results: For each dataset, we sample a 30-second validation clip and show keyframes every 2 seconds. Rows: predicted signs from the segmentation model (yellow), original subtitles (green), SEA-aligned subtitles (blue), and expert-aligned ground truth (purple). In general, segmentation identifies the signing frames of interest; SEA then shifts subtitles toward higher text–sign similarity—for example, in BSL the first subtitle begins at the fingerspelling AMPNIBIS (``amphibians'') and ends at the sign for ``sliminess''.}
    \label{fig:qualitative_results}
\end{figure*}

This section introduces the datasets (\S\ref{sec:datasets}), our evaluation protocol, and the baselines (\S\ref{sec:baseline}). 
We then present how our SEA method performs both quantitatively (\S\ref{sec:results}), compared to previous SOTA approaches, and qualitatively (\S\ref{sec:qualitative}).
Finally, we ablate different components of our approach (\S\ref{sec:ablation}).

\subsection{Dataset Overview}
\label{sec:datasets}

We select continuous sign language datasets that include \textit{parallel} subtitles before and after manual alignment, such as BOBSL, How2Sign, and WMT-SLT SRF. We present an overview of the statistics of the datasets in Table \ref{tab:datasets}.
For BOBSL and How2Sign, we use the default split.
For WMT-SLT SRF, we re-split the original training set of the WMT-SLT 22 campaign to obtain 4 episodes for dev and test, respectively, as the subtitles before manual alignment are not available for these sets.

We additionally hire DSGS experts to annotate (annotation details in Appendix \ref{sec:human_annotation}) 14 episodes of the \textit{mitenand} program from the SwissSLi dataset, which acts as a test-only set in a highly low-resource setting.
This dataset is intended to be a DSGS/German dataset of superior quality, as it was translated by deaf translators offline rather than live interpreted, unlike WMT-SLT SRF, where the hearing interpreters tend to revert to spoken-language grammar and drop words under time pressure.

\subsection{Evaluation and Baselines}
\label{sec:baseline}

We follow the evaluation protocol proposed by \citet{bull2021aligning}, but prioritize the \textit{F1-score} over frame accuracy, since SEA does not predict at frame level. 
Specifically, \textit{F1@0.50} counts a subtitle as correct when the predicted segment overlaps the ground truth with intersection-over-union (IoU) $\geq 0.50$.

We construct two baselines per dataset: 
\textit{Original} refers to the original subtitles from the datasets, typically intended to align with the speech signal; \textit{Original\textsuperscript{+}} refers to adding a fixed offset to the start and end position of each subtitle\footnote{Equivalent to $S_{\text{audio}}$ and $S_{\text{audio}}^{+}$ in \citet{bull2021aligning}.}. 
The fixed offset numbers are set to the median differences between the default and the manually aligned subtitles in the training set\footnote{Unavailable for SwissSLi, validation set is used instead.}, as collected in Table \ref{tab:datasets}.
For BOBSL, we keep +2.7 seconds — the \emph{average} shift between audio and manually-aligned sentences in the manual training set, as specified by prior work \cite{dataset:albanie2021bobsl}. However, we find that using median values is more robust to outliers for other datasets.

For BOBSL, we have two additional baselines: \textit{SAT} \citep{bull2021aligning} and  \textit{SAT\textsuperscript{+}} \citep{jang2025deep}. We also train the SAT model on the WMT-SLT data to test the applicability of SAT on a different dataset, using the same I3D visual backbone but domain-specific alignment supervision.

\subsection{Experimental Results and Discussion}
\label{sec:results}

Table \ref{tab:results} reports SEA’s performance: it attains SOTA on all four test sets—surpassing SAT/SAT\textsuperscript{+} on BOBSL and establishing new SOTA on three additional, cross-language/domain datasets not previously evaluated. Key points follow.

\paragraph{The original alignment quality varies across datasets.}
Fixed offsets significantly improve interpreted datasets (e.g., BOBSL, WMT-SLT SRF), where a typical interpretation delay is expected; however, the effect is weaker for datasets with different production workflows.
Note that alignment on How2Sign is inherently hard because some random appearances of signing do not correspond to any subtitle, especially near episode starts and ends.

\paragraph{End-to-end training by SAT/SAT\textsuperscript{+} remains a good choice when supervision is available.}
The effectiveness also relies on a strong in-domain visual backbone network (I3D for BOBSL), which does not necessarily excel on another dataset or sign language, as seen in the case of WMT-SLT. 
We do not further explore SAT\textsuperscript{+} in this work and refer to additional optimizations by \citet{jang2025deep}.

\paragraph{Adjusting original alignment based on the prosodic cues from segmentation already helps.} 
In \textit{Segment and Align}, the segmentation plays two main roles:
(1) detecting in which temporal spans signing happens (against the signer making a pause/rest or being absent in the picture);
(2) segmenting meaningful units to draw decision boundaries on, such as a prosodic pause.
However, the lack of semantic understanding limits performance, especially when such cues are rare, e.g., How2Sign, where signers tend to sign continuously without many breaks in a studio recording setting. 

\paragraph{The embedding module further enhances the ability to discern boundaries and achieves SOTA.}
Consistent improvements across datasets and languages are observed after the \textit{embed} step is added to the pipeline, even with the \textit{SignCLIP-multilingual} model, which did not see any DSGS/German data during the pretraining phase.
Language-specific finetuning further enhances the alignment quality: 
For BOBSL, finetuning on in-domain sign spottings, although noisy, brings a boost of $\sim$4 to 6 \textit{F1@0.50} on the validation and test sets.
For How2Sign, finetuning on three out-of-domain but large-scale ISLR datasets also clearly helps.
The effect of finetuning on Signsuisse, a small lexicon resource of a different domain, is relatively weak as seen for WMT-SLT and SwissSLi.

\paragraph{Prior alignment and dataset biases matter.} 
For BOBSL, we additionally initialize the prior location of the subtitles by the model output from \citet{jang2025deep}, as a controlled experiment to explore SEA’s sensitivity and robustness to prior locations of different quality.
When initialized with the subtitles automatically aligned by \textit{SAT\textsuperscript{+}}, SEA inherits the gains and further refines them.
When initialized with subtitles in a challenging original position, such as in How2Sign, SEA still improves alignment but less markedly.
It is also worth mentioning the importance of dataset-specific biases. In the random search process, we have learned similar offset values as in the \textit{Original\textsuperscript{+}} baseline and found that a fixed 1-second offset applied after the DP process is always useful in the datasets evaluated\footnote{``Useful'' in the sense of better evaluation results since annotators tend to give extra duration after the signing stops.}.

\subsection{Qualitative Results}
\label{sec:qualitative}


We present qualitative results in Figure \ref{fig:qualitative_results}, which include a 30-second sample for each of the evaluated datasets, taken from SEA's global alignment.
In general, the sign segmentation accurately identifies temporal regions of interest and excludes non-signing parts, such as pauses during signing. 
The segmentation and embedding modules, together, help the final alignment step move the original subtitles to locations closer to the human-aligned ground truth. 
However, because SEA still tends to stay tied to the original subtitle timings, when the latter are poor, it affects the final alignment quality, which can sometimes cause errors to propagate. 
In our ASL example, the first subtitle misses several signs—including TENT—which is tricky since the word ``tent'' also appears in the next sentence.

\subsection{Ablation Study}
\label{sec:ablation}

We ablate the effects of different segmentation and embedding methods in Table \ref{tab:bsl-segmentation} and Table \ref{tab:islr-align}.



\paragraph{Segmentation.} We explore different segmentation models introduced in \S\ref{sec:implementation} and measure their impact on the \textit{Segment and Align} approach. We evaluate segmentation performance on the BSL Corpus (BSLCP), equipped with human-timed gloss annotations, and alignment on the BOBSL validation set using the same segmentation model. 
We cannot evaluate segmentation on BOBSL since it lacks equivalent gloss-level human annotation with accurate timings. 
\citet{Renz2021signsegmentation_a,Renz2021signsegmentation_b} performs better than \citet{moryossef-etal-2023-linguistically} on segmentation evaluation, but the final alignment is poorer.
We speculate three reasons:
(1) BSL Corpus might not be representative of BOBSL due to the domain gap.
(2) \citet{Renz2021signsegmentation_a,Renz2021signsegmentation_b}, trained to excel at discerning signing boundaries, produce many false positives of signing when used for alignment.
(3) Segmentation of linguistically well-defined signs might not be a hard requirement for downstream tasks such as alignment. Upon inspection, \citet{moryossef-etal-2023-linguistically} produces slightly over-segmented output, but SEA seems to operate well on the \textit{sub-word} \cite{sennrich-etal-2016-neural} level.
Finetuning \citet{moryossef-etal-2023-linguistically} on the BSLCP training data achieves SOTA segmentation performance (even outperforming follow-up work; \citet{he2025hands}), but falls short of the original model's alignment capabilities (\textit{Segmentation-BSLCP} in Table \ref{tab:results}). 
We thus adhere to the original segmentation model from \citet{moryossef-etal-2023-linguistically} for SEA.

\paragraph{Embedding.} We also examine the dynamics between the strength of the embedding models, measured by retrieval-based ISLR, and the final alignment results. 
For BSL, we test ISLR on the BOBSL sign spottings, and for ASL, we test ISLR on the PopSign ASL dataset. A consistent positive correlation is observed between ISLR and alignment across languages.
For BOBSL, we further test a specific gloss-based alternative of the embedding module, i.e., using the pseudo/human glosses produced in the previous CSLR2 \cite{raude2024tale} work.
We fill in the similarity matrix with a value of 1 if a sign segment's gloss is in a subtitle, and 0 otherwise.
This alternative can be viewed as a rigid assignment of signs (analogous to how a human expert would approach it, as shown in Figure \ref{fig:teaser}) rather than a soft similarity; we find that the latter approach works better. 

\begin{table}[ht]
    \centering
    \resizebox{\linewidth}{!}{%
        \begin{tabular}{l|cc|c}
        \toprule
        \textbf{Model} & \multicolumn{2}{c|}{\textbf{BSLCP Seg.}} & \textbf{BOBSL Align.} \\
        \cmidrule(lr){2-3}\cmidrule(lr){4-4}
        & \textbf{F1} & \textbf{mF1S} & \textbf{F1@50} \\
        \midrule

        \citet{Renz2021signsegmentation_a} & 47.71 & -- & 57.98 \\
        \citet{Renz2021signsegmentation_b} & -- & -- & 57.53 \\
        \citet{he2025hands} & 50.18 & -- & -- \\

        \midrule
        
        \citet{moryossef-etal-2023-linguistically} & 31.13 & 33.32 & \textbf{66.24} \\

        \midrule
        
        finetuned on BSLCP & 51.46 & 55.46 & -- \\
        finetuned + negative sampling & \textbf{52.09} & \textbf{56.04} & 63.61 \\

        \bottomrule
        \end{tabular}
    }
    \caption{Sign segmentation results on BSL Corpus test set and alignment results on BOBSL validation set using the same segmentation model in \textit{Segment and Align}.}
    \label{tab:bsl-segmentation}
\end{table}

\begin{table}[ht]
    \centering
    \resizebox{\linewidth}{!}{%
        \begin{tabular}{l|l|cc}
        \toprule
        \textbf{Lang.} & \textbf{Model} & \textbf{ISLR} & \textbf{BOBSL Align.} \\
        \midrule

        \multirow[c]{4}{*}[-0.3ex]{\makecell[l]{\textbf{BSL}\\[0.2ex]{\footnotesize (BOBSL)}}} 
        & SignCLIP-multilingual & \phantom{0}0.5 & 66.70 \\
        & SignCLIP-BSL & \textbf{43.0} & \textbf{72.78} \\

        \cmidrule(lr){2-4}
        
        & CSLR2 pseudo glosses & -- & 68.80 \\
        & CSLR2 human glosses* & -- & 78.75 \\

        \midrule

        \multirow{2}{*}{\shortstack[l]{\textbf{ASL}\\[0.4ex]{\footnotesize (PopSign)}}}
& SignCLIP-multilingual & \phantom{0}3.0 & 35.51 \\
& SignCLIP-ASL & \textbf{84.3} & \textbf{38.32} \\

        \bottomrule
        \end{tabular}
    }
    \caption{ISLR results on BOBSL sign spottings and PopSign ASL test sets with respect to alignment scores on validation sets in \textit{Segment, Embed, and Align}. Results using CSLR2 glosses are also included as an ablation, and the human glosses* are marked as an oracle.}
    \label{tab:islr-align}
\end{table}




\section{Conclusion}

We introduce \textit{Segment, Embed, and Align} (SEA), a universal method for aligning subtitles to sign language videos. 
Splitting the problem into sign \emph{segmentation}, cross-modal \emph{embedding}, and global \emph{alignment} via DP, SEA attains SOTA alignment results across diverse languages and data sources. 
The modular design of SEA enables plug-and-play improvements—language-specific finetuning strengthens embeddings without requiring end-to-end alignment training on in-domain supervision.

Beyond establishing strong empirical results, SEA has practical value in solving this specific problem: it can harvest higher-quality parallel text–sign data at scale and serve as a flexible component in broader sign language processing pipelines. 
We release all components of SEA to encourage reproducible research and future advances.

\section*{Limitations}
\label{sec:limit}

Our work has the following limitations:
(1) The embedding and alignment models can be possibly improved iteratively with regard to each other, and we leave this to future work due to time and space limits.
(2) Although already evaluated extensively on four diverse datasets, the robustness to dataset-specific biases and data quality can be further improved. For example, a mechanism for the algorithm to drop irrelevant signing or subtitles would alleviate the bad cases observed in How2Sign, and merging segments when the boundary is difficult to discern would improve accuracy and be useful for paragraph-level downstream applications.
(3) Human-in-the-loop workflows, such as post-editing and evaluation, are not studied in this work, but we believe that it will be valuable to establish a semi-automatic process with human annotators' intervention to further enhance alignment quality and practical usability.


\section*{Acknowledgments}
The BOBSL images in this paper are used with the kind permission of the BBC.
We also thank SWISS TXT and Amit Moryossef for their help in organizing the DSGS annotation campaign.
We thank the DSGS expert annotators for their efforts in aligning the 14 episodes of the \textit{mitenand} program.

ZJ is funded by the Swiss Innovation Agency (Innosuisse) flagship IICT (PFFS-21-47). This work was supported by the ANR project
CorVis ANR-21-CE23-0003-01, the UKRI EPSRC Programme Grant SignGPT EP/Z535370/1, and a Royal Society
Research Professorship RSRP$\backslash$R$\backslash$241003,
the Institute of Information \& communications Technology Planning \& Evaluation (IITP) grant funded by the Korean government (MSIT, RS-2025-02263977, Development of Communication Platform supporting User Anonymization and Finger Spelling-Based Input Interface for Protecting the Privacy of Deaf Individuals).

We thank the VGG and Oxford colleagues for supporting ZJ's research visit and work, especially Sarah Clayton, for the welcome, and Ashish Thandavan, for maintaining the compute cluster.

\bibliography{shortstrings, custom}

\clearpage
\newpage
\appendix

\section{Extended Implementation Details}
\label{sec:extended_details}

In this section, we provide an additional explanation of the alignment algorithm in SEA with Python-like pseudocode (Figure \ref{fig:sea_pseudocode}).
Following \S\ref{sec:method}, we first obtain segmented signs (line 2), and then embed these signs together with the subtitles (line 5-6).

\paragraph{Similarity matrix.} The subtitle and signing embeddings form a similarity matrix \( \mathbb{R}^{N_{\text{subtitles}} \times N_{\text{signs}}} \), where each entry is the dot product between a subtitle and a sign embedding, measuring their similarity (line 7). 
We explore two variants for assigning groups of consecutive signs to subtitle units, without/with the use of a similarity matrix. The subtitles are then shifted according to the matched signs. 
Following \citet{bull2021aligning}, alignment adjusts subtitle timings to match the signing. 
Unlike frame-level classification in local windows, SEA globally assigns groups of consecutive signs to subtitle units and shifts subtitles accordingly.

\paragraph{Segment and Align.} A \textit{cost function} is defined for assigning a \texttt{span} of signs to a subtitle \texttt{cue} (line 12). 
The cost is a weighted sum of four terms: (a) distance between subtitle start and first sign; (b) distance between subtitle end and last sign; (c) duration difference between subtitle and span; (d) internal gaps within the span. 
By minimizing this cost globally, subtitles are aligned to signing using only prosodic cues and prior subtitle timing, referred to as \textit{Segment and Align}.

\begin{figure}[ht]
\centering
\begin{minipage}{\linewidth}
\begin{lstlisting}[
  style=clipstyle,
  numbers=left,
  numberstyle=\scriptsize\color{gray},
  numbersep=6pt,
  % Reserve space so numbers sit inside the box
  xleftmargin=1.3em,
  % frame=single,
  framexleftmargin=1.3em
]
# §3.1 Segment
signs = segmenter(video)

# §3.2 Embed
v_emb = embed_sign(signs) 
t_emb = embed_text(subtitles) 
sim_matrix = dot_product(t_emb, v_emb)

# §3.3 Align
# Define alignment cost function 
# between a subtitle cue and a span of signs
def cost(cue, span):
  a = abs(cue.start - span.start)
  b = abs(cue.end - span.end)
  c = abs(cue.duration - span.duration)
  d = compute_internal_gap(span)
  e = compute_total_similarity(cue, span)
  return a + b + w_dur*c + w_gap*d - w_sim*e

# Filter and normalize similarity scores
sim_matrix = zero_out(sim_matrix, win_size)
sim_matrix = softmax_normalize(sim_matrix)

# Run dynamic programming
dp = aligner(signs,subtitles,cost,sim_matrix)
bounds = backtrack(dp)

# Postprocessing: 
# refine span with best subgroup
for i in range(len(subtitles)):
  group = signs[bounds[i]:bounds[i+1]]
  subgroups = split_on_gap(group, max_gap)
  alignments[i] = \\
    pick_best(subgroups, subtitles[i], cost)

# Finally, update subtitle timestamps
for i in range(len(subtitles)):
  subtitles[i].start = alignments[i][0].start
  subtitles[i].end   = alignments[i][-1].end
\end{lstlisting}
\end{minipage}
\caption{Python-like pseudocode for SEA: segment, embed, and align using DP with a weighted cost function, followed by refining spans and updating subtitles.}
\label{fig:sea_pseudocode}
\end{figure}

\paragraph{Segment, Embed, and Align.}
To incorporate semantic associations between text and signing, we add a negative cost term, (e) \texttt{total\_similarity}, which sums the subtitle–sign similarities from the matrix. 
With this addition, the process becomes \textit{Segment, Embed, and Align}. 

Since episodes can be almost 1 hour long and only nearby signs are assumed \textit{alignable} with each subtitle, we restrict locality with a \texttt{win\_size} (window size) parameter, zeroing out scores outside this window of closest neighbors (line 21). 
Remaining scores are row-normalized with \texttt{Softmax}, yielding a matrix of values concentrated around the temporal diagonal (line 22).
We employ a dynamic programming algorithm to determine the global optimum (line 25). 
A \texttt{max\_gap} constraint further picks the most similar sign subgroup for each subtitle with gaps no larger than this threshold, regulating the maximum gap of signs allowed within a subtitle, and discarding the rest (line 30).
Finally, the subtitles' start/end times are updated (line 37).

\paragraph{DP implementation.}
We maintain a table \(dp[i,j]\), of the best total cost after aligning the first \(i\) subtitles to the first \(j\) signs:
\[
\textstyle dp[i,j]=\min_{i-1\le k<j}\{\,dp[i-1,k]+\phi(t_i,s_{k{:}j-1})\,\}
\]
\(dp[0,0]=0\), \(dp[0,j>0]=\infty\).
After filling the table iteratively via the recurrence above, we do \emph{not} require including all signs since false positives are often found in the episode end; instead we pick the best ending \(j^\star=\arg\min_{0\le j\le m} dp[n,j]\) and backtrack from \((n,j^\star)\) to \((0,0)\) by the stored predecessors to recover the subtitle–sign assignment.

\paragraph{Parameter random search.}

\begin{table*}[ht]
\centering
\resizebox{\linewidth}{!}{%
\sisetup{round-mode=places, round-precision=2}
\begin{tabular}{l *{4}{S[table-format=2.2]}}
\toprule
\textbf{Parameter} &
\multicolumn{1}{c}{\textbf{BOBSL}} &
\multicolumn{1}{c}{\textbf{How2Sign}} &
\multicolumn{1}{c}{\textbf{WMT\text{-}SLT}} &
\multicolumn{1}{c}{\textbf{SwissSLi}} \\
\midrule
Start offset before DP                & 2.6  & 1.3  & 1.4  & 0.49 \\
End offset before DP                  & 2.1  & 1.5  & 1.6  & 0.56 \\
Start offset after DP                 & 0.0  & 0.0  & 0    & 0    \\
End offset after DP                   & 1.0  & 1.0  & 1    & 1    \\
\texttt{b-threshold} for segmentation & 30   & 40   & 20   & 20   \\
\texttt{o-threshold} for segmentation & 50   & 50   & 30   & 30   \\
Window size (\texttt{win\_size})                          & 50   & 50   & 50   & 50   \\
Max inter-sign gap (\texttt{max\_gap})                     & 10   & 8    & 6    & 6    \\
Duration distance weight (\texttt{w\_dur})              & 1    & 5    & 0.5  & 0.5  \\
Inter-sign gap weight (\texttt{w\_gap})                & 5    & 0.8  & 5    & 5    \\
Similarity weight (\texttt{w\_sim})                  & 10   & 10   & 5    & 1    \\
\bottomrule
\end{tabular}
}
\caption{Best parameter settings per dataset (offsets and gaps in seconds).}
\label{tab:parameters}
\end{table*}

\begin{figure*}
    \centering
    \includegraphics[width=\linewidth]{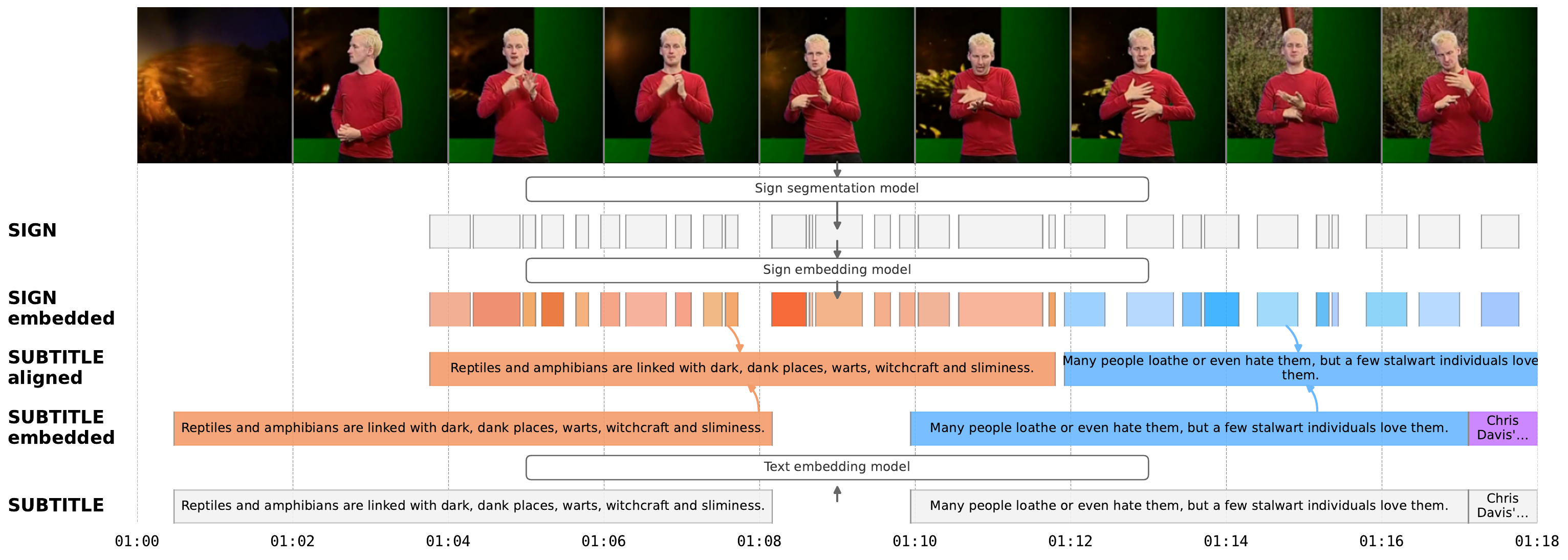}
    \caption{\textbf{Aligning subtitles to continuous signing using SEA:} From top down, signs are first segmented, then embedded (colored). From bottom up, subtitles are embedded into the same latent space as signs and then aligned accordingly to the direction of similarities.}
    \label{fig:method_detail}
\end{figure*}

For each of the datasets evaluated, we first conduct a random parameter search of up to \num{10000} iterations (approximately 1 day, running on 128 CPU workers) to find the optimal parameter combination under the condition of \textit{Segment and Align}.
We then inherit the best parameter combination when introducing the \textit{Embed} step, only slightly adjusting the weight for similarity—\texttt{w\_sim}—with the rest unchanged, as searching again with the embedding step is significantly more costly.

The values of these parameters are presented in Table \ref{tab:parameters}.
We note that these parameters might reflect dataset-specific biases learned from the training set.
For SwissSLi, which lacks a training set, we transfer most parameters from WMT-SLT, a dataset of the same languages.
The similarity weights for both are set relatively small, as the corresponding embedding models are relatively weakly finetuned on a small amount of data in DSGS/German.

\section{SEA Alignment Illustrated on BOBSL}

In addition to the description in \S\ref{sec:extended_details}, Figure \ref{fig:method_detail} provides a more visual, step-by-step illustration of how SEA alignment works using the example video with id \texttt{5224144816887051284} from BOBSL.
Signs (from top down) and subtitles (from bottom up) are aligned based on their similarities in the embedding space.

\section{Human Annotation Details on SwissSLi}
\label{sec:human_annotation}

We hire DSGS experts to annotate 14 episodes of the \textit{mitenand} program from the SwissSLi dataset \cite{jiang-etal-2024-swisssli}.
The main task is to manually align the subtitles of \textit{mitenand} in German to better match the signing in DSGS.

The annotators are either DSGS interpreter students or deaf interpreters.
For the three students, annotations were performed as part of a course project in exchange for 3 ECTS course credits. Participation was voluntary.
For deaf interpreters, the annotation was compensated at the rate of $\sim$40 USD per hour with informed consent.

We append the annotation instructions below:

\paragraph{Annotation instructions.} The videos we are using are from the mitenand in Gebärdensprache program. They are translated by deaf translators. You are asked to correct the annotation of the videos by ELAN.

You can find the videos and ELAN files in SWITCHdrive. After you are finished, upload the new ELAN files to the same location.

When you open an ELAN file with the corresponding video under the same folder, you will see four existing annotation tiers:

SUBTITLE: the German sentences from the subtitle/caption file, whose timings might not be perfectly aligned to the signing.
SENTENCE: empty sentences detected by our AI model, which might also be inaccurate, but could serve as an indication for correcting the timings of the subtitles.
SIGN: empty signs detected by our AI model, which might also be inaccurate, but could serve as an initial proposal for individual sign annotations.
GLOSS: signs in the SIGN tier filled automatically with the glosses from the gloss files the deaf translators use. Note that the number of actual signs might vary from the detected ones.

You have two tasks:

On the sentence level, copy the SUBTITLE tier and create a new SUBTITLE\_CORRECTED tier, where the start and end times of each sentence are corrected to match the signing.
On the sign level, copy the GLOSS tier and create a new GLOSS\_CORRECTED tier, where the start and end times of each gloss are corrected to match the signing, and the gloss name is corrected if the current one is wrong (just use intuitive glosses, they will be changed to match our DSGS lexical database later). In addition, remove superfluous gloss annotation as well as segments such as “Mit Untertiteln von SWISS TXT” and add missing gloss annotation if needed. If necessary, split segments (e.g., split “SAGEN NEIN” into “SAGEN” and “NEIN”). If lexical meaning is expressed non-manually, please use a gloss in square brackets; e.g., if the meaning of “no” is expressed with a headshake, please create a segment “[NO]”. Please apply narrow segmentation following the description from this article: “Transitional movements between signs do not count as part of either sign”. Therefore, usually, there are gaps between two signs during which the articulators move from the end of one sign to the beginning of the next.”

It would be great if you could also note a few things during the process:

How much time does it take you to finish each task? What is the most time-consuming part?
Does the output of the AI model help you with the process, for example, the SENTENCE tier on sentence timing correction and the automatically generated glosses on gloss annotation?
Any issue with the model’s automatic annotation or the glosses from the deaf translators?

\end{document}